\documentclass[runningheads]{llncs}
\usepackage{graphicx}
\usepackage{tikz}
\usepackage{comment}
\usepackage{amsmath,amssymb} % define this before the line numbering.
\usepackage{color}
\usepackage{mathrsfs}
\usepackage{graphicx}
\usepackage{subfigure}
\usepackage{marvosym}

\usepackage[accsupp]{axessibility}  % Improves PDF readability for those with disabilities.

\newif\ifcomment
\commentfalse% \questiontwotrue

\ifcomment
\usepackage{ruler}
\usepackage[width=122mm,left=12mm,paperwidth=146mm,height=193mm,top=12mm,paperheight=217mm]{geometry}
\fi

\begin{document}
% \linenumbers
\pagestyle{headings}
\mainmatter
\def\ECCVSubNumber{11}  % Insert your submission number here

\title{Learning an Efficient Multimodal Depth Completion Model} % Replace with your title

%\begin{comment}
\ifcomment
\titlerunning{ECCV-22 submission ID \ECCVSubNumber}
\authorrunning{ECCV-22 submission ID \ECCVSubNumber}
\author{Anonymous ECCV submission}
\institute{Paper ID \ECCVSubNumber}
\fi
%\end{comment}
%******************

\titlerunning{Efficient Multimodal Depth Completion}
% \titlerunning{EMDC}
\authorrunning{D. Hou et al.}

\author{
Dewang Hou$^{1}$, Yuanyuan Du$^{2}$, Kai Zhao$^{3}$, Yang Zhao$^{4,}$\textsuperscript{{\Letter}} \\
$^1$Peking University, China\\
$^2$Chongqing University, China\\
$^3$Tsinghua University, China\\
$^4$Hefei University of Technology, China\\
{\tt\small dewh@pku.edu.cn; dyy$\_$0805@126.com; zhaok18@tsinghua.org.cn; yzhao@hfut.edu.cn}
}

\institute{ }
\maketitle
\let\thefootnote\relax\footnotetext{
\Letter \; Corresponding author.
}

\begin{comment}
\titlerunning{Abbreviated paper title}
\author{First Author\inst{1}\orcidID{0000-1111-2222-3333} \and
Second Author\inst{2,3}\orcidID{1111-2222-3333-4444} \and
Third Author\inst{3}\orcidID{2222--3333-4444-5555}}
%
\authorrunning{F. Author et al.}
\institute{Princeton University, Princeton NJ 08544, USA \and
Springer Heidelberg, Tiergartenstr. 17, 69121 Heidelberg, Germany
\email{lncs@springer.com}\\
\url{http://www.springer.com/gp/computer-science/lncs} \and
ABC Institute, Rupert-Karls-University Heidelberg, Heidelberg, Germany\\
\email{\{abc,lncs\}@uni-heidelberg.de}}
\end{comment}
%******************

\begin{abstract}
With the wide application of sparse ToF sensors in mobile devices, RGB image-guided sparse depth completion has attracted extensive attention recently, but still faces some problems. First, the fusion of multimodal information requires more network modules to process different modalities. But the application scenarios of sparse ToF measurements usually demand lightweight structure and low computational cost. Second, fusing sparse and noisy depth data with dense pixel-wise RGB data may introduce artifacts. In this paper, a light but efficient depth completion network is proposed, which consists of a two-branch global and local depth prediction module and a funnel convolutional spatial propagation network. The two-branch structure extracts and fuses cross-modal features with lightweight backbones. The improved spatial propagation module can refine the completed depth map gradually. Furthermore, corrected gradient loss is presented for the depth completion problem. Experimental results demonstrate the proposed method can outperform some state-of-the-art methods with a lightweight architecture. The proposed method also wins the championship in the MIPI2022 RGB+TOF depth completion challenge.

\keywords{depth completion, sparse ToF, RGB guidance}
\end{abstract}

\section{Introduction}

Depth information plays an important role in various vision applications, such as autonomous driving, robotics, augmented reality, and 3D mapping. In past decades, many depth sensors have been developed and applied to obtain depth information, such as time-of-flight (ToF) and light detection and ranging (LiDAR) sensors. With the rapid development of smart mobile devices, sparse ToF depth measurements have attracted extensive attention recently due to their unique advantages. For example, sparse ToF sensors can effectively avoid multi-path interference problem that usually appears in full-field ToF depth measurements. In addition, the power consumption of sparse ToF depth measurement is much lower than that of full-field ToF sensors, which is very suitable for mobile devices and edge devices to reduce battery consumption and heating. However, there is a proverb that says you can't have your cake and eat it too. The sparse ToF depth measurement also has significant disadvantages. First, sparse ToF cannot provide a dense pixel-wise depth map due to hardware limitations such as the small amount of laser pulse. Second, sparse ToF depth data easily suffers from noise. To overcome these limitations, depth completion (DC), which converts a sparse depth measurement to a dense pixel-wise depth map, becomes very useful due to industrial demands.

Single modal sparse depth completion is a quite difficult and ill-posed problem because dense pixel-wise depth values are severely missing. Fortunately, aligned RGB images usually can be obtained in these practical scenarios. Therefore, recent depth completion methods often utilize RGB guidance to boost depth completion and show significantly better performance than the unguided way. The straightforward reason is that natural images provide plentiful semantic cues and pixel-wise details, which are critical for filling missing values in sparse depth modality.

For traditional full-field ToF depth maps, many RGB image guided depth super-resolution or enhancement methods have been proposed, e.g., traditional learning-based joint enhancement models~\cite{yang2014color,kwon2015data}, and recent deep neural network (DNN)-based joint depth image restoration methods~\cite{ma2018sparse,cheng2018depth,yang2019dense,tang2020learning,park2020non}. For the sparse ToF depth measurements, Ma et al.~\cite{ma2018sparse} proposed a single depth regression network to estimate dense scene depth by means of RGB guidance. Tang et al.~\cite{tang2020learning} presented a learnable guided convolution module and applied it to fuse cross-modality information between sparse depth and corresponding image contents.

These RGB-guided depth completion methods can produce better depth maps by comparing them with single modal depth completion. However, current depth completion methods with RGB guidance still face the following two challenges. First, multi-modality usually leads to more computational cost, because different network modules are required for these modalities with different distributions. Note that sparse depth sensors are usually applied in mobile devices or other edge devices with finite computing power and real-time requirement. As a result, the depth completion models not only need to extract, map and fuse multimodal information effectively but also need to meet lightweight architecture and real-time demand. Second, different to full-field depth maps with pixel-wise details, the sparse depth data may introduce extra artifacts during the fusion process. Moreover, the noise contained in sparse depth data may further intensify the negative effects~\label{problems}.

In this paper, we proposed a light but efficient multimodal depth completion network based on the following three aspects: fusing multi-modality data more effectively; reducing the negative effects of missing values regions in sparse modality; recovering finer geometric structures for both objective metrics and subjective quality. 
Overall, the contributions are summarized as follows:
\textbf{1)} We present a lightweight baseline, which can achieve state-of-the-art results while being computationally efficient. 
More specifically, the two-branch framework is first used to estimate global and local predictions. Features are exchanged between these two branches to fully make use of cross-modal information. A fusion module with relative confidence maps is presented after two branches, which can further optimize the fused results and accelerate convergence. 
\textbf{2)} we propose a funnel convolutional spatial propagation network, which can gradually improve the depth features with reweighting strategy. 
\textbf{3)} In addition, corrected gradient loss is designed for depth completion problems correspondingly. Experimental results show that the proposed method can achieve superior performance, but also run very fast in the inference stage.

The remainder of this paper is organized as follows. Related works are briefly reviewed in Section 2. Section 3 introduces the proposed method in detail. Experimental results are analyzed in Section 4, and Section 5 concludes the paper.

\section{Related Work}

\subsection{Unguided Depth Completion}
Unguided DC methods tend to estimate dense depth map from a sparse depth map directly. Uhrig et al.~\cite{01uhrig2017sparsity} first applied a sparsity invariant convolutional neural network (CNN) for DC task. Thereafter, many DC networks have been proposed by using the strong learning capability of CNNs~\cite{02eldesokey2018propagating,03eldesokey2020uncertainty}. Moreover, some unguided DC methods~\cite{05lu2020depth,07lu2022depth} use RGB images for auxiliary training, but still perform unguided DC in the inference stage. Unguided methods don’t rely on corresponding RGB images, but their performance is naturally worse than RGB-guided DC methods~\cite{hu2022deep}. Because RGB guidance can provide more semantic information and pixel-wise local details. This paper focuses on sparser ToF depth data in real-world applications, and thus mainly discusses the related RGB-guided DC methods.

% % 04, 11, 22, 25, 33, 34, 36
\subsection{RGB-guided Depth Completion}
Currently, different types of guided DC methods have been proposed to employ corresponding RGB images to improve completed depth maps. These RGB-guided DC methods can be roughly divided into four categories, i.e., early fusion model, late fusion model, explicit 3D representation model, and spatial propagation model.

Early fusion methods~\cite{ma2018sparse,12senushkin2021decoder,13dimitrievski2018learning,14hambarde2020s2dnet,15imran2019depth,16xu2019depth} directly concatenate depth map and RGB image or their shallow features before feeding them into the main body of networks. These methods usually adopt encoder-decoder structure~\cite{11ma2018sparse,12senushkin2021decoder} or two-stage coarse to refinement prediction~\cite{13dimitrievski2018learning,14hambarde2020s2dnet}. Note that the coarse-to-refinement structure is also frequently used in subsequent residual depth map prediction models~\cite{17liao2017parse,18gu2021denselidar}.
Compared to early fusion, late fusion model uses two sub-network to extract features from RGB image and input depth data, and then fuses cross-modal information in the immediate or late stage. Since two modalities are used, this type of method usually adopts dual-branch architecture, such as dual-encoder with one decoder~\cite{19jaritz2018sparse,20fu2020depth,21zhong2019deep}, dual-encoder-decoder~\cite{tang2020learning,23schuster2021ssgp,24yan2021rignet}, and global and local depth prediction~\cite{van2019sparse,26lee2020deep}.
In addition, there are some 3D representation models learn 3D representations to guide depth completion, such as 3D-aware convolution~\cite{28chen2019learning,29zhao2021adaptive} and surface normal representation method~\cite{30qiu2019deeplidar}.

Spatial propagation network (SPN)~\cite{31iu2017learning} is also a typical model widely used in DC tasks by treating DC as a depth regression problem. Cheng et al.~\cite{32cheng2019learning} first applied SPN to the DC task by proposing a convolutional SPN (CSPN). Then, they presented CSPN++~\cite{cheng2020cspn++} by means of context-aware and resource-aware CSPN. To take advantage of both SPN and late fusion models, Hu et al.~\cite{hu2021penet} further proposed a precise and efficient DC network (PENet). Subsequently, deformable SPN (DSPN)~\cite{35xu2020deformable} and attention-based dynamic SPN (DySPN)~\cite{lin2022dynamic} were also applied for RGB-guided DC problem, and achieved impressive performance on KITTI depth completion benchmark~\cite{01uhrig2017sparsity}.

\section{Methods}

\begin{figure*}[htbp]
\centering
\includegraphics[width=0.88\textwidth]{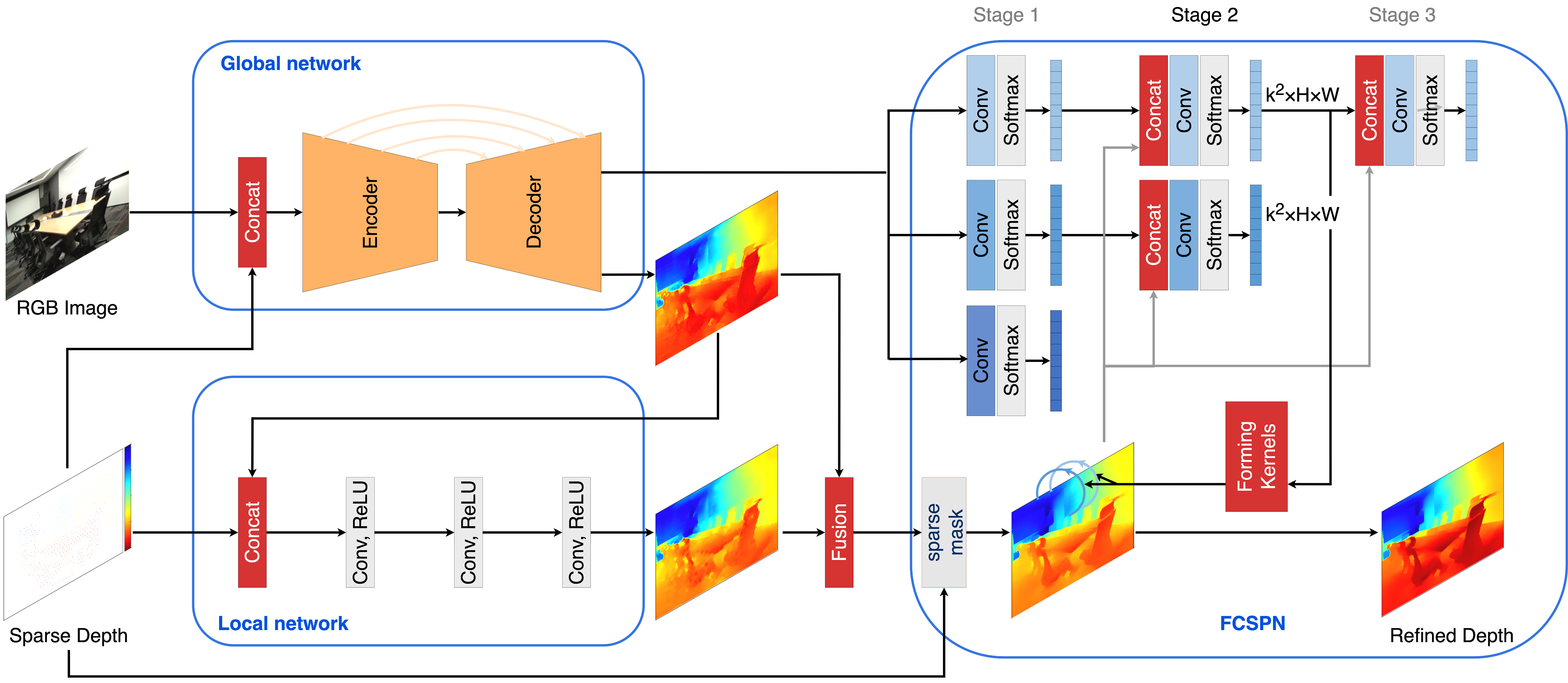}

\caption{Network architecture of the proposed EMDC. We introduce a pipeline that consists of two-branch global and local network, fusion module and FCSPN. As an illustrative example, we show the second stage in which FCSPN generates two convolution kernels.}
\label{fig_EMDC}
\end{figure*}

In this paper, we thoroughly study the key components of depth completion methods to derive an Efficient Multimodal Depth Completion network (EMDC), shown in Fig.~\ref{fig_EMDC}, which tends to solve the above mentioned problems. In the following, we introduce each module of EMDC in detail.

% \ifcomment
% \fi

\subsection{Global and Local Depth Prediction with Fusion}

\begin{figure*}[htbp]
\centering
\includegraphics[width=0.88\textwidth]{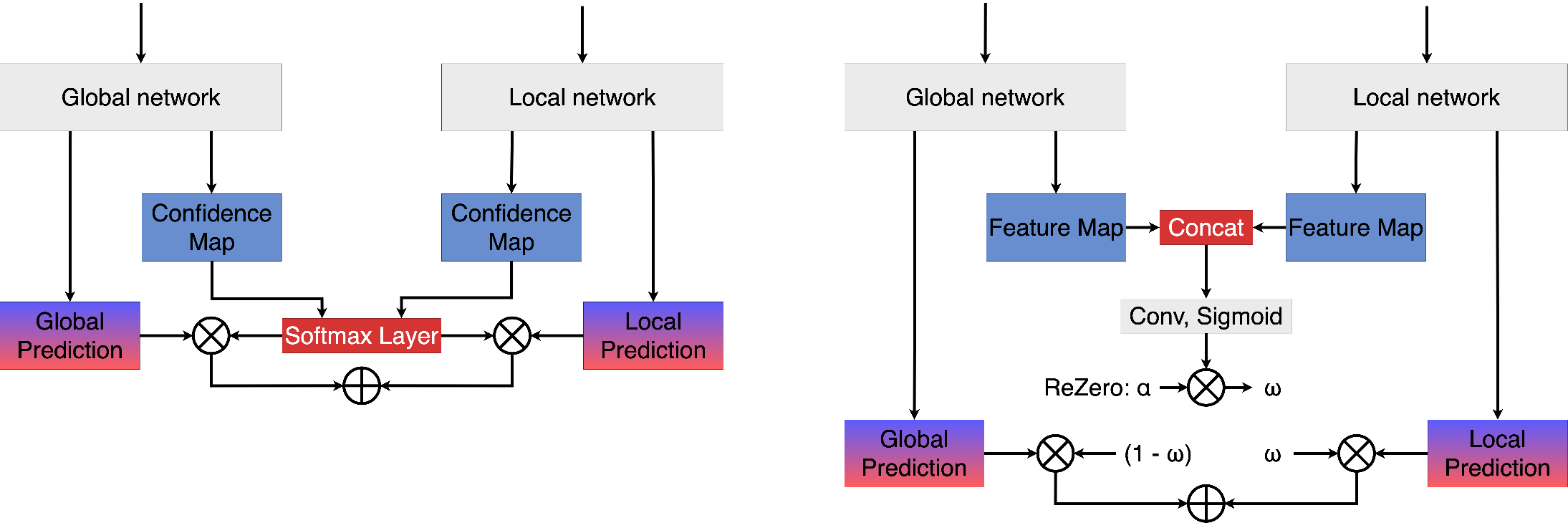}

\caption{Left: fusion module proposed in~\cite{van2019sparse}, Right: our proposed fusion module.}
\label{fig_ReZeroFusion}
\end{figure*}

Motivated by~\cite{hu2022deep,van2019sparse,26lee2020deep}, a global and local depth prediction (GLDP) framework is adopted to fuse the information from two modalities efficiently.
 As illustrated in Fig.~\ref{fig_EMDC}, different architectures are designed for the global and local depth estimation sub-networks, respectively. For the global branch, which needs to estimate depth data with the help of RGB image, a typical U-Net structure with large respective fields is used. In addition, pretrained MobileNetV2~\cite{sandler2018mobilenetv2} is adopted as the encoder in the global branch to obtain strong learning capability with lightweight structures. For the local branch that processes simple depth prediction and homogeneous sparse depth, we merely design a lightweight CNN module to reproduce depth map. Finally, the output depth maps of these two branches are fused via a relative fusion module. For the confidence maps used by the fusion module, we think they should reflect the relative certainty of global and local depth predictions, the so-called relative confidence maps. As shown in Fig.~\ref{fig_ReZeroFusion}, we adjust the pathways of the features to involve this idea directly. Furthermore, we improve the relative fusion module by zero-initializing the relative confidence maps of local depth prediction with the zero-initialization method~\cite{bachlechner2021rezero}, which can achieve faster convergence and better test performance in our experiments. Related experimental results can be found from the ablation results in Table~\ref{tab_ablation_mipi}.

To reduce the negative effects introduced by the distribution characteristics of sparse modality, we use pixel-shuffle (PS)~\cite{shi2016real} to replace the traditional upsampling layer in the global branch. Compared to normal upsampling, PS operation doesn’t magnify the non-ideal missing information. In addition, the batch normalization layers in the local branch are all removed, as they are fragile to features with anisotropic spatial distribution and slightly degrade the results.

\subsection{Funnel Convolutional Spatial Propagation Network}

The most important target of the DC task is to recover better scene structures for both objective metrics and subjective quality. Therefore, to alleviate the problem of recovering scene structures, we propose a Funnel Convolutional Spatial Propagation Network (FCSPN) for depth refinement. As shown in Fig.~\ref{fig_EMDC}, the FCSPN inherits the basic structure of the spatial propagation network (SPN)~\cite{cheng2018depth,cheng2020cspn++,park2020non,lin2022dynamic,hu2021penet}, which is a popular refinement technique in the DC field. FCSPN can fuse the point-wise results from large to small dilated convolutions in each stage. The maximum dilation of the convolution at each stage is reduced gradually, thus forming a funnel-like structure stage by stage. Compared with CSPN++~\cite{cheng2020cspn++}, the FCSPN generates a new set of kernels, termed reweighting, at each stage to increase the representation capability of SPNs. The FCSPN embraces dynamic filters with kernel reweighting which has a more substantial adaptive capability than adjusting them via attention mechanism~\cite{lin2022dynamic}. Moreover, the kernel reweighting mechanism merely relies on the kernels in the previous stage and the current depth map in the refinement process, thus reducing the computation complexity considerably. Experimental results show that FCSPN not only has superior performance, but also runs fast in inference, as shown in Table~\ref{table:runtime}.

\subsection{Loss Functions}

 \begin{figure}[htbp]
	\centering
	\subfigure[ground truth depth]{
		\begin{minipage}[b]{0.3\linewidth}
			\includegraphics[width=1\linewidth]{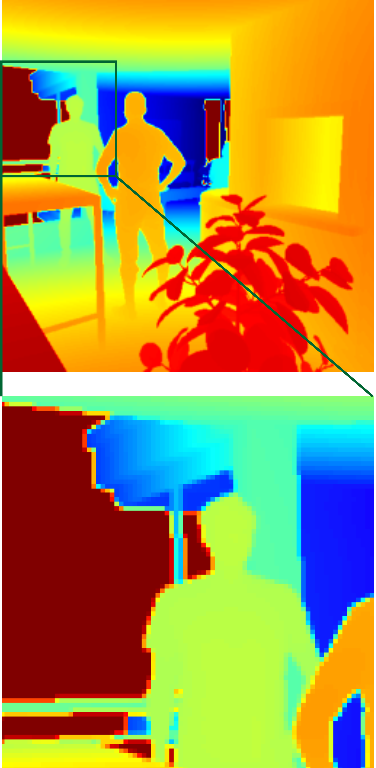}
	\end{minipage}}
	\subfigure[gradient map]{
		\begin{minipage}[b]{0.3\linewidth}
			\includegraphics[width=1\linewidth]{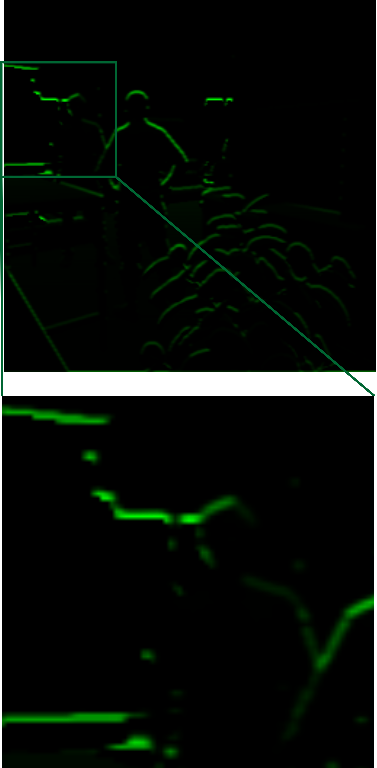}
	\end{minipage}}
	\subfigure[corrected gradient map]{
		\begin{minipage}[b]{0.3\linewidth}
			\includegraphics[width=1\linewidth]{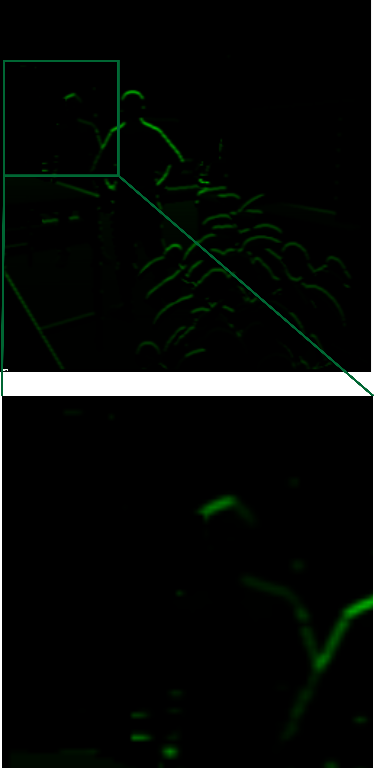}
	\end{minipage}}
	\caption{Gradient maps obtained by filtering a depth map. And for simplicity, we only show normalized y-gradient from Sobel-Feldman operator in the vertical direction. And (b) is a naive gradient map and (c) is produced by our method. }
	\label{fig_GDL}
\end{figure}

Sophisticated depth completion algorithms often focused on minimizing pixel-wise errors which typically leads to over-smoothed results~\cite{wang2009mean}.
Degradation of structural information would impede their practical applications, such as safety issues in autonomous driving.
To handle this problem, we proposed a corrected gradient loss for the depth completion task. Gradient loss~\cite{mathieu2015deep} was originally proposed to preserve the structure of natural images. Intuitively, applying it for depth map regression problems may also help to refine the blurry predictions obtained from MAE or MSE losses.
However, it is difficult to obtain a complete precisely annotated depth dataset. One of the problems brought about by this is that the gradients, such as those extracted with the Sobel filtering, between depth points that are unreachable by depth measurement and other depth points are inaccurate. We elaborate on this phenomenon in Fig.~\ref{fig_GDL}.
The dark red regions in the distance in Fig.~\ref{fig_GDL} are actually unreachable by depth measurement. It can be clearly seen that the ``annotated" depth of these regions is all set to zero, so the gradient between other areas and them is also unreliable, and even reversed. To handle this problem, we propose the corrected gradient loss. We formulate it as follows:
\begin{eqnarray}
	\mathcal{L}_{cgdl}(\hat{Y},Y)
	&=& \left \| F(\hat{Y}) - F(Y) \right \|_{p} * E(sgn(Y \in \nu  )),
\label{eq:lossgdl}
\end{eqnarray}
where $F(\cdot)$ denotes a specific gradient operator or set of gradient operators. $\nu$ denotes the valid value range for depth measurement. $sgn(\cdot)$ denotes signum function that extracts the sign of its inputs. $E(\cdot)$ denotes the erosion operation, which is a morphology-based image processing method.
$E(sgn(Y \in \nu  )$ can also be understood as playing the role of online data cleaning.
The loss $\mathcal{L}_{cgdl}$ can be easily introduced to the existing DC models and the final loss function in our work is:
\begin{eqnarray}
	\mathcal{L}(\hat{Y},Y)
	& = \mathcal{L}_{l1}(\hat{Y},Y) + {\lambda}_{1} * \mathcal{L}_{l1}(\hat{Y}_{global},Y) \\
	& + {\lambda}_{2} * \mathcal{L}_{l1}(\hat{Y}_{local},Y) + {\lambda}_{3} * \mathcal{L}_{cgdl}(\hat{Y},Y),
\label{eq:lossall}
\end{eqnarray}
where
\begin{equation}
	{\lambda}_{1} = \mathcal{L}_{l1}(\hat{Y},Y) / \mathcal{L}_{l1}(\hat{Y}_{global},Y),
\label{eq:loss1}
\end{equation}
\begin{equation}
	{\lambda}_{2} = \mathcal{L}_{l1}(\hat{Y},Y) / \mathcal{L}_{l1}(\hat{Y}_{local},Y),
\label{eq:loss2}
\end{equation}
\begin{equation}
	{\lambda}_{3} = 0.7 * \mathcal{L}_{l1}(\hat{Y},Y) / \mathcal{L}_{cgdl}(\hat{Y},Y)
\label{eq:loss3}
\end{equation}
We also utilize an adaptive loss weight adjustment strategy to search for optimum fusion in the training process. Its implementation can be seen from equations~\ref{eq:loss1} and equations~\ref{eq:loss2}. The losses calculated on the results of the two-branch network are strictly constructed to be the same value.
Because these losses should be optimized synchronously rather than alternately to match the fusion strategy in the GLDP framework and avoid mode collapse caused by model mismatch~\cite{goodfellow2014generative}.

\section{Experiments}

\subsection{Datasets and Metrics}

Before Mobile Intelligent Photography and Imaging (MIPI) challenge, there lacks a high-quality sparse depth completion dataset, which hindered related research for both industry and academia. To this end, the RGBD Challenge of MIPI provided a high-quality synthetic DC dataset (MIPI dataset), which contains 20,000 pairs of pre-aligned RGB and ground truth depth images of 7 indoor scenes. For each scene, the RGB and the ground-truth depth are rendered along a smooth trajectory in a created 3D virtual environment.

For testing, the testing data in MIPI dataset comes from mixed sources, including synthetic data, spot-iToF, and some samples subsampled from dToF in iPhone.
These mixed testing data can verify the robustness and generalization ability of different DC methods in real-world scenarios.

Four metrics are employed to evaluate the performance of depth completion algorithms, i.e., Relative Mean Absolute Error (RMAE), Edge Weighted Mean Absolute Error (EWMAE), Relative Depth Shift (RDS), and Relative Temporal Standard Deviation (RTSD). Details of these metrics can be found in the technical reports of the MIPI RGBD challenge.
The final score combines these four metrics as follows:
\begin{equation}
    {score = 1 - 1.8 \times {R\!M\!A\!E} - 0.6 \times {E\!W\!M\!A\!E} - 3 \times {R\!D\!S} - 4.6 \times {R\!T\!S\!D}}
\label{eq:score}
\end{equation}

\subsection{Implementational Details}
In training stage, we use the AdamW optimizer~\cite{loshchilov2017decoupled} with ${\beta}_{1}$=0.9, ${\beta}_{2}$=0.999. We set the initial learning rate as 0.001, and use a learning scheduler of cosine annealing with a warm start.
After a linear increase in the first 10 epochs, the learning rate gradually decreases along the cosine annealing curve after each mini-batch within one period.
The proposed EMDC is trained with a mini-batch size of 10 in a total of 150 epochs.
The training samples are further augmented via random flip and color jitter. In addition, we use the exponential moving average (EMA) technique to obtain a more stable model.
Our algorithm is implemented in PyTorch. A full training of our network takes about 12 hours on an NVIDIA Tesla A6000 GPU. The source codes are available along with the pretrained models to facilitate reproduction.$^{1}$
\footnote{$^{1}$\url{https://github.com/dwHou/EMDC-PyTorch}}

\begin{figure}[htbp]
	\centering
	\subfigure[RGB guidance]{
		\begin{minipage}[b]{0.295\linewidth}
			\includegraphics[width=1\linewidth]{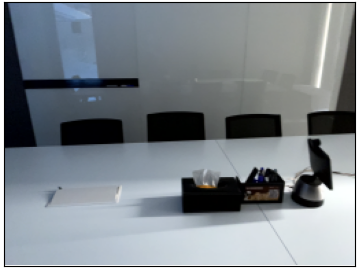}
	\end{minipage}}
	\subfigure[sparse depth map]{
		\begin{minipage}[b]{0.295\linewidth}
			\includegraphics[width=1\linewidth]{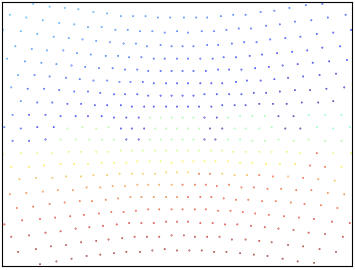}
	\end{minipage}}
	\subfigure[ground truth]{
		\begin{minipage}[b]{0.295\linewidth}
			\includegraphics[width=1\linewidth]{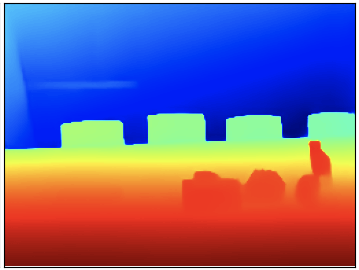}
	\end{minipage}}

	\subfigure[FusionNet]{
		\begin{minipage}[b]{0.45\linewidth}
			\includegraphics[width=1\linewidth]{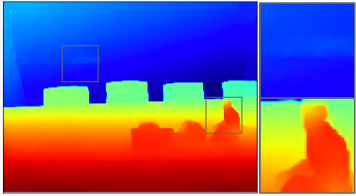}
	\end{minipage}}
	\subfigure[CSPN]{
		\begin{minipage}[b]{0.45\linewidth}
			\includegraphics[width=1\linewidth]{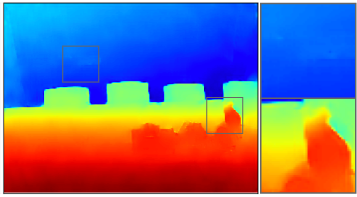}
	\end{minipage}}
	\subfigure[PENet]{
		\begin{minipage}[b]{0.45\linewidth}
			\includegraphics[width=1\linewidth]{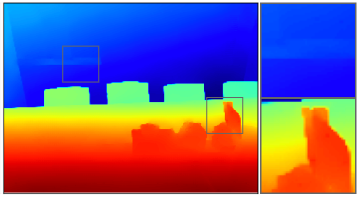}
	\end{minipage}}
	\subfigure[EMDC]{
		\begin{minipage}[b]{0.45\linewidth}
			\includegraphics[width=1\linewidth]{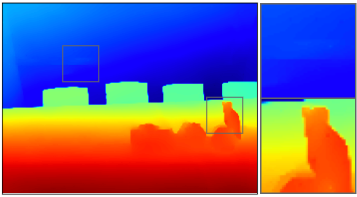}
	\end{minipage}}
	\caption{Depth completion results of different methods. }

\end{figure}	

\begin{figure}[t]
	\centering
\subfigure[RGB guidance]{
		\begin{minipage}[b]{0.295\linewidth}
			\includegraphics[width=1\linewidth]{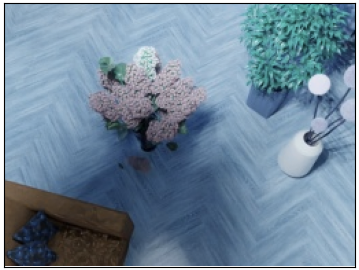}
	\end{minipage}}
	\subfigure[sparse depth map]{
		\begin{minipage}[b]{0.295\linewidth}
			\includegraphics[width=1\linewidth]{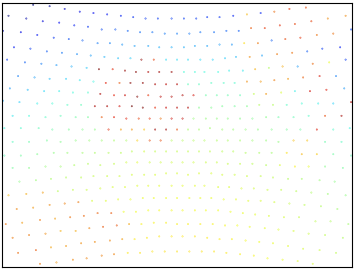}
	\end{minipage}}
	\subfigure[ground truth]{
		\begin{minipage}[b]{0.295\linewidth}
			\includegraphics[width=1\linewidth]{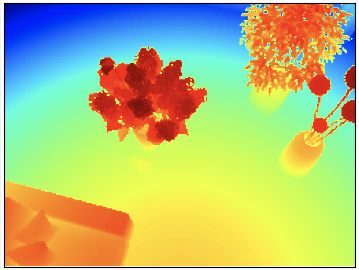}
	\end{minipage}}
	
	\subfigure[FusionNet]{
		\begin{minipage}[b]{0.45\linewidth}
			\includegraphics[width=1\linewidth]{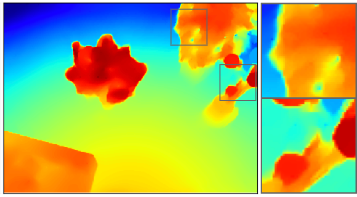}
	\end{minipage}}
	\subfigure[CSPN]{
		\begin{minipage}[b]{0.45\linewidth}
			\includegraphics[width=1\linewidth]{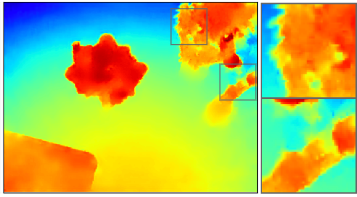}
	\end{minipage}}
	\subfigure[PENet]{
		\begin{minipage}[b]{0.45\linewidth}
			\includegraphics[width=1\linewidth]{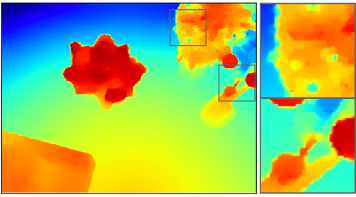}
	\end{minipage}}
	\subfigure[EMDC]{
		\begin{minipage}[b]{0.45\linewidth}
			\includegraphics[width=1\linewidth]{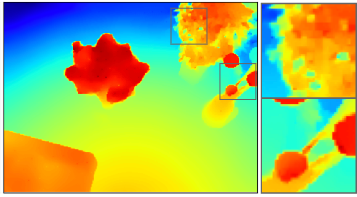}
	\end{minipage}}

	\caption{Different depth completion results of another indoor scene.}
	
\end{figure}

\ifcomment
\begin{figure}[htbp]
	\centering
	\begin{minipage}{0.49\linewidth}
		\centering
		\includegraphics[width=0.9\linewidth]{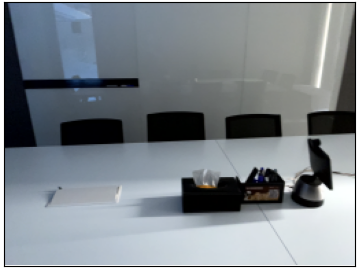}
		\caption{chutian1}
		\label{chutian1}%文中引用该图片代号
	\end{minipage}
	\begin{minipage}{0.49\linewidth}
		\centering
		\includegraphics[width=0.9\linewidth]{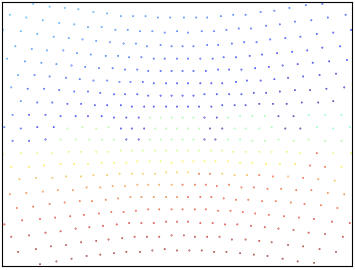}
		\caption{chutian2}
		\label{chutian2}%文中引用该图片代号
	\end{minipage}
	%\qquad
	%让图片换行，
	
	\begin{minipage}{0.3\linewidth}
		\centering
		\includegraphics[width=0.85\linewidth]{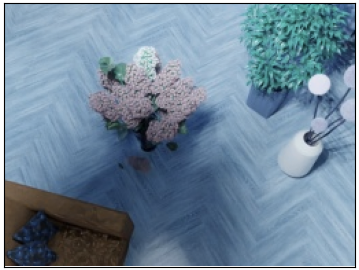}
		\caption{chutian3}
		\label{chutian3}%文中引用该图片代号
	\end{minipage}
	\begin{minipage}{0.3\linewidth}
		\centering
		\includegraphics[width=0.85\linewidth]{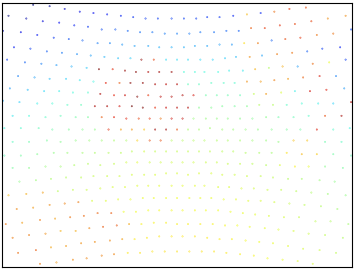}
		\caption{chutian4}
		\label{chutian4}%文中引用该图片代号
	\end{minipage}
	\begin{minipage}{0.3\linewidth}
		\centering
		\includegraphics[width=0.85\linewidth]{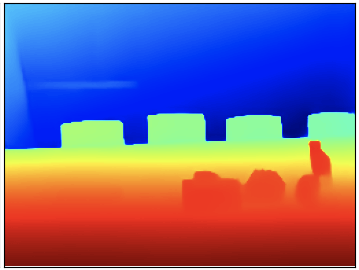}
		\caption{chutian4}
		\label{chutian4}%文中引用该图片代号
	\end{minipage}
	\begin{minipage}{0.45\linewidth}
		\centering
		\includegraphics[width=0.9\linewidth]{Used/FusionNet1.png}
		\caption{chutian4}
		\label{chutian4}%文中引用该图片代号
	\end{minipage}
	\begin{minipage}{0.45\linewidth}
		\centering
		\includegraphics[width=0.9\linewidth]{Used/CSPN1.png}
		\caption{chutian4}
		\label{chutian4}%文中引用该图片代号
	\end{minipage}
	\begin{minipage}{0.45\linewidth}
		\centering
		\includegraphics[width=0.9\linewidth]{Used/PENet1.png}
		\caption{chutian4}
		\label{chutian4}%文中引用该图片代号
	\end{minipage}
	\begin{minipage}{0.45\linewidth}
		\centering
		\includegraphics[width=0.9\linewidth]{Used/Ours1.png}
		\caption{chutian4}
		\label{chutian4}%文中引用该图片代号
	\end{minipage}	
\end{figure}
\fi

\subsection{Experimental Results}

\begin{table}[htbp]
\begin{tabular}{clcccccllccccccc}
\hline
\textbf{Method} \quad &  & \multicolumn{1}{l}{} & \quad Overall Score↑ \quad&  &  & \quad RMAE↓ \quad  & \multicolumn{1}{c}{} & \multicolumn{1}{c}{} & \quad EWMAE↓ \quad &  &  & \quad RDS↓ \quad  &  &  & \quad RTSD↓ \quad \\ \hline
FusionNet~\cite{van2019sparse}       &  &                      & 0.795  &  &  & 0.019 &                      &                      & 0.094  &  &  & 0.009 &  &  & 0.019 \\
CSPN~\cite{32cheng2019learning}            &  &                      & 0.811  &  &  & 0.015 &                      &                      & 0.090  &  &  & 0.007 &  &  & 0.019 \\
PENet~\cite{hu2021penet}           &  &                      & 0.840  &  &  & 0.014 &                      &                      & 0.087  &  &  & 0.003 &  &  & \textcolor{blue}{0.016} \\ \hline

2rd in MIPI      &  &                      & \textcolor{blue}{0.846}  &  &  & \textcolor{red}{0.009} &                      &                      & \textcolor{blue}{0.085}  &  &  & \textcolor{blue}{0.002} &  &  & 0.017 \\
3rd in MIPI      &  &                      & 0.842  &  &  & \textcolor{blue}{0.012} &                      &                      & 0.087  &  &  & \textcolor{red}{0.000} &  &  & 0.018\\
\hline
EMDC (ours)     &  &                      & \textcolor{red}{0.855}  &  &  & \textcolor{blue}{0.012} &                      &                      & \textcolor{red}{0.084}  &  &  & \textcolor{blue}{0.002} &  &  & \textcolor{red}{0.015} \\
\hline
\end{tabular}
\caption{Objective results of different DC methods on MIPI dataset. Red/Blue text: best/second-best.}
\label{tab_results mipi}
\end{table}

Table~\ref{tab_results mipi} lists the experimental results of different DC methods, i.e., CSPN~\cite{32cheng2019learning}, PENet~\cite{hu2021penet}, FusionNet~\cite{van2019sparse}, and the proposed EMDC.
From Table~\ref{tab_results mipi}, we can find that the proposed EMDC can achieve better performance than efficient CSPN, PENet and FusionNet.
In addition, Table~\ref{tab_results mipi} also lists the performance of SOTA methods of other top teams. Owing to the well-designed architectures, the proposed EMDC can produce better results on EWMAE, RTSD and overall scores.

Fig. 4 and Fig. 5 illustrate some subjective results of different DC methods. Overall, the proposed EMDC can reproduce shaper edges, smoother flat surfaces, and clearer details. As shown in Fig. 4, the EMDC can reproduce a more complete flat surface, but other methods cannot well handle different colors on the same surface. By comparing the tiny lines in Fig. 5, the proposed method can reproduce better details than other methods.

\begin{table}[htbp]
	\centering
	\footnotesize
	\begin{tabular}{l|l|l|l|l|l|l|l|l}
		\hline
		Method            & (a) & (b) & (c) & (d) & (e) & (f) & (g) & (h) \\
		\hline
		\hline
		FCSPN$_{s=6}$    & & \checkmark & & & & &  \\
		\hline
		FCSPN$_{s=9}$    & & & \checkmark & \checkmark & \checkmark & \checkmark & \checkmark & \checkmark   \\
		\hline
		Pixel-Shuffle$_{~global~network}$  & & & & \checkmark & \checkmark & \checkmark & \checkmark & \checkmark \\
		\hline
		Remove BN$_{~local~network}$  & & & & & \checkmark & \checkmark & \checkmark & \checkmark \\
		\hline
		Relative Certainty  & & & & & & \checkmark & \checkmark & \checkmark  \\
		\hline
		ReZero Init   & & & & & & & \checkmark & \checkmark \\
		\hline
		Corrected GDL & \checkmark & \checkmark & \checkmark & \checkmark & \checkmark & \checkmark &  & \checkmark  \\
		\hline
		\hline
		MIPI overall score$~\uparrow$   & 0.798 & 0.819 & 0.826 & 0.834 & 0.838 & 0.849 & 0.847 & 0.855 \\
		\hline
	\end{tabular}
	\caption{Ablation study on the proposed methods. The results are evaluated by the scoring method mentioned in section 4.4.}
	\label{tab_ablation_mipi}
\end{table}

\begin{table}[htbp]
	\begin{center}
		\begin{tabular}{c|c|c|c}
		\hline
			Method & SPN Part & Iterations & Propagation Time \\ \hline
			UNet+CSPN~\cite{cheng2018depth} & CSPN  & 21 & 25.3ms \\
			EMDC & FCSPN$_{s=6}$  & 15 & 15.7ms \\
			EMDC & FCSPN$_{s=9}$  & 21 & 18.2ms \\ \hline
			%B & \textemdash &0.016s \\
			%B+GE & \textemdash &0.017s \\
		\end{tabular}
	\end{center}
	\caption{Runtime of CSPN and the proposed EMDC on 256$\times$192 resolution.}
	\label{table:runtime}
\end{table}

\noindent{\bfseries Ablation Studies.}
For ablation testing, we gradually change various settings of the proposed EMDC. All ablation tests are conducted under the same experiment setup, and results are reported in Table~\ref{tab_ablation_mipi}.
Ablation results in Table~\ref{tab_ablation_mipi} demonstrate the effectiveness of these modules or settings. The relative certainty fusion module and corrected gradient loss can significantly improve the performance. Using Pixel-Shuffle layer, removing BN layers, and applying re-zero initialization are also helpful to sparse DC tasks. In addition, FCSPN with 9 stages outperforms that with 6 stages, but more iterations also lead to more time-cost.

\noindent{\bfseries Inference Time.}
As sparse ToF sensors are mainly applied in power-limited mobile devices nowadays,
the power consumption and processing time are important factors in DC tasks.
Hence, we compared the running time of CSPN and FCSPN on the same NVIDIA Tesla V100 GPU.
In Table~\ref{table:runtime}, EMDC(FCSPN$_{s=6}$) consists of 6 stages with a total of 15 iterations, and EMDC(FCSPN$_{s=9}$) consists of 9 stages with 21 iterations.
The improved FCSPN structure can run faster than efficient CSPN.
Note that, by comparing the overall scores, although the number of propagation of EMDC(FCSPN$_{s=6}$) is less than CSPN, it can perform better than CSPN on the MIPI Dataset.

\section{Conclusions}

This paper proposed a lightweight and efficient multimodal depth completion (EMDC) model for sparse depth completion tasks with RGB image guidance. The EMDC consists of a two-branch global and local depth prediction (GLDP) module and a funnel convolutional spatial propagation network (FCSPN). 
The global branch and local branch in GLDP extract depth features from pixel-wise RGB image and sparse depth data according to the characteristics of different modalities.
An improved fusion module with relative confidence maps is then applied to fuse multimodal information. The FCSPN module can refine both the objective and subjective quality of depth maps gradually. A corrected gradient loss is also presented for the depth completion problem. Experimental results show the effectiveness of the proposed method. In addition, the proposed method wins the first place in the RGB+ToF Depth Completion track in Mobile Intelligent Photography and Imaging (MIPI) challenge.

\clearpage
\bibliographystyle{splncs04}
\bibliography{egbib}

\begin{thebibliography}{10}
\providecommand{\url}[1]{\texttt{#1}}
\providecommand{\urlprefix}{URL }
\providecommand{\doi}[1]{https://doi.org/#1}

\bibitem{bachlechner2021rezero}
Bachlechner, T., Majumder, B.P., Mao, H., Cottrell, G., McAuley, J.: Rezero is
  all you need: {Fast} convergence at large depth. In: Uncertainty in
  Artificial Intelligence. pp. 1352--1361. PMLR (2021)

\bibitem{28chen2019learning}
Chen, Y., Yang, B., Liang, M., Urtasun, R.: Learning joint {2D-3D}
  representations for depth completion. In: Proceedings of the IEEE/CVF
  International Conference on Computer Vision (ICCV). pp. 10023--10032 (2019)

\bibitem{cheng2020cspn++}
Cheng, X., Wang, P., Guan, C., Yang, R.: {CSPN++}: Learning context and
  resource aware convolutional spatial propagation networks for depth
  completion. In: Proceedings of the AAAI Conference on Artificial Intelligence
  (AAAI). vol.~34, pp. 10615--10622 (2020)

\bibitem{cheng2018depth}
Cheng, X., Wang, P., Yang, R.: Depth estimation via affinity learned with
  convolutional spatial propagation network. In: Proceedings of the European
  Conference on Computer Vision (ECCV). pp. 103--119 (2018)

\bibitem{32cheng2019learning}
Cheng, X., Wang, P., Yang, R.: Learning depth with convolutional spatial
  propagation network. IEEE Transactions on Pattern Analysis and Machine
  Intelligence  \textbf{42}(10),  2361--2379 (2019)

\bibitem{13dimitrievski2018learning}
Dimitrievski, M., Veelaert, P., Philips, W.: Learning morphological operators
  for depth completion. In: International Conference on Advanced Concepts for
  Intelligent Vision Systems (ACIVS). pp. 450--461. Springer (2018)

\bibitem{03eldesokey2020uncertainty}
Eldesokey, A., Felsberg, M., Holmquist, K., Persson, M.: Uncertainty-aware
  {CNNs} for depth completion: Uncertainty from beginning to end. In:
  Proceedings of the IEEE/CVF Conference on Computer Vision and Pattern
  Recognition (CVPR). pp. 12014--12023 (2020)

\bibitem{02eldesokey2018propagating}
Eldesokey, A., Felsberg, M., Khan, F.S.: {Propagating confidences through CNNs
  for sparse data regression}. arXiv preprint arXiv:1805.11913  (2018)

\bibitem{20fu2020depth}
Fu, C., Dong, C., Mertz, C., Dolan, J.M.: Depth completion via inductive fusion
  of planar {LiDAR} and monocular camera. In: IEEE/RSJ International Conference
  on Intelligent Robots and Systems (IROS). pp. 10843--10848 (2020)

\bibitem{goodfellow2014generative}
Goodfellow, I., Pouget-Abadie, J., Mirza, M., Xu, B., Warde-Farley, D., Ozair,
  S., Courville, A., Bengio, Y.: Generative adversarial nets. Proceedings on
  the International Conference on Neural Information Processing Systems (NIPS)
  \textbf{27} (2014)

\bibitem{18gu2021denselidar}
Gu, J., Xiang, Z., Ye, Y., Wang, L.: Dense{LiDAR}: A real-time pseudo dense
  depth guided depth completion network. IEEE Robotics and Automation Letters
  \textbf{6}(2),  1808--1815 (2021)

\bibitem{14hambarde2020s2dnet}
Hambarde, P., Murala, S.: S2dnet: Depth estimation from single image and sparse
  samples. IEEE Transactions on Computational Imaging  \textbf{6},  806--817
  (2020)

\bibitem{hu2022deep}
Hu, J., Bao, C., Ozay, M., Fan, C., Gao, Q., Liu, H., Lam, T.L.: {Deep Depth
  Completion: A Survey}. arXiv preprint arXiv:2205.05335  (2022)

\bibitem{hu2021penet}
Hu, M., Wang, S., Li, B., Ning, S., Fan, L., Gong, X.: {PENet}: Towards precise
  and efficient image guided depth completion. In: IEEE International
  Conference on Robotics and Automation (ICRA). pp. 13656--13662 (2021)

\bibitem{15imran2019depth}
Imran, S., Long, Y., Liu, X., Morris, D.: Depth coefficients for depth
  completion. In: Proceedings of the IEEE/CVF Conference on Computer Vision and
  Pattern Recognition (CVPR). pp. 12438--12447 (2019)

\bibitem{19jaritz2018sparse}
Jaritz, M., De~Charette, R., Wirbel, E., Perrotton, X., Nashashibi, F.: Sparse
  and dense data with {CNNs}: {Depth completion} and semantic segmentation. In:
  IEEE International Conference on 3D Vision (3DV). pp. 52--60 (2018)

\bibitem{kwon2015data}
Kwon, H., Tai, Y.W., Lin, S.: Data-driven depth map refinement via multi-scale
  sparse representation. In: Proceedings of the IEEE/CVF Conference on Computer
  Vision and Pattern Recognition (CVPR). pp. 159--167 (2015)

\bibitem{26lee2020deep}
Lee, S., Lee, J., Kim, D., Kim, J.: Deep architecture with cross guidance
  between single image and sparse {LiDAR} data for depth completion. IEEE
  Access  \textbf{8},  79801--79810 (2020)

\bibitem{17liao2017parse}
Liao, Y., Huang, L., Wang, Y., Kodagoda, S., Yu, Y., Liu, Y.: Parse geometry
  from a line: Monocular depth estimation with partial laser observation. In:
  IEEE International Conference on Robotics and Automation (ICRA). pp.
  5059--5066 (2017)

\bibitem{lin2022dynamic}
Lin, Y., Cheng, T., Zhong, Q., Zhou, W., Yang, H.: Dynamic spatial propagation
  network for depth completion. arXiv preprint arXiv:2202.09769  (2022)

\bibitem{31iu2017learning}
Liu, S., De~Mello, S., Gu, J., Zhong, G., Yang, M.H., Kautz, J.: Learning
  affinity via spatial propagation networks. Proceedings on the International
  Conference on Neural Information Processing Systems (NIPS)  \textbf{30}
  (2017)

\bibitem{loshchilov2017decoupled}
Loshchilov, I., Hutter, F.: Decoupled weight decay regularization. arXiv
  preprint arXiv:1711.05101  (2017)

\bibitem{05lu2020depth}
Lu, K., Barnes, N., Anwar, S., Zheng, L.: From depth what can you see? depth
  completion via auxiliary image reconstruction. In: Proceedings of the
  IEEE/CVF Conference on Computer Vision and Pattern Recognition (CVPR). pp.
  11306--11315 (2020)

\bibitem{07lu2022depth}
Lu, K., Barnes, N., Anwar, S., Zheng, L.: Depth completion auto-encoder. In:
  IEEE/CVF Winter Conference on Applications of Computer Vision Workshops
  (WACVW). pp. 63--73 (2022)

\bibitem{ma2018sparse}
Ma, F., Karaman, S.: {Sparse-to-Dense}: Depth prediction from sparse depth
  samples and a single image. In: IEEE International Conference on Robotics and
  Automation (ICRA). pp. 4796--4803 (2018)

\bibitem{11ma2018sparse}
Ma, F., Karaman, S.: {Sparse-to-Dense}: Depth prediction from sparse depth
  samples and a single image. In: IEEE International Conference on Robotics and
  Automation (ICRA). pp. 4796--4803 (2018)

\bibitem{mathieu2015deep}
Mathieu, M., Couprie, C., LeCun, Y.: Deep multi-scale video prediction beyond
  mean square error. arXiv preprint arXiv:1511.05440  (2015)

\bibitem{park2020non}
Park, J., Joo, K., Hu, Z., Liu, C.K., So~Kweon, I.: Non-local spatial
  propagation network for depth completion. In: Proceedings of the European
  Conference on Computer Vision (ECCV). pp. 120--136. Springer (2020)

\bibitem{30qiu2019deeplidar}
Qiu, J., Cui, Z., Zhang, Y., Zhang, X., Liu, S., Zeng, B., Pollefeys, M.:
  Deep{LiDAR}: Deep surface normal guided depth prediction for outdoor scene
  from sparse {LiDAR} data and single color image. In: Proceedings of the
  IEEE/CVF Conference on Computer Vision and Pattern Recognition (CVPR). pp.
  3313--3322 (2019)

\bibitem{sandler2018mobilenetv2}
Sandler, M., Howard, A., Zhu, M., Zhmoginov, A., Chen, L.C.: {MobileNetV2}:
  Inverted residuals and linear bottlenecks. In: Proceedings of the IEEE/CVF
  Conference on Computer Vision and Pattern Recognition (CVPR). pp. 4510--4520
  (2018)

\bibitem{23schuster2021ssgp}
Schuster, R., Wasenmuller, O., Unger, C., Stricker, D.: {SSGP}: Sparse spatial
  guided propagation for robust and generic interpolation. In: Proceedings of
  the IEEE/CVF Winter Conference on Applications of Computer Vision (WACV). pp.
  197--206 (2021)

\bibitem{12senushkin2021decoder}
Senushkin, D., Romanov, M., Belikov, I., Patakin, N., Konushin, A.: Decoder
  modulation for indoor depth completion. In: IEEE/RSJ International Conference
  on Intelligent Robots and Systems (IROS). pp. 2181--2188 (2021)

\bibitem{shi2016real}
Shi, W., Caballero, J., Husz{\'a}r, F., Totz, J., Aitken, A.P., Bishop, R.,
  Rueckert, D., Wang, Z.: Real-time single image and video super-resolution
  using an efficient sub-pixel convolutional neural network. In: Proceedings of
  the IEEE/CVF Conference on Computer Vision and Pattern Recognition (CVPR).
  pp. 1874--1883 (2016)

\bibitem{tang2020learning}
Tang, J., Tian, F.P., Feng, W., Li, J., Tan, P.: Learning guided convolutional
  network for depth completion. IEEE Transactions on Image Processing
  \textbf{30},  1116--1129 (2020)

\bibitem{01uhrig2017sparsity}
Uhrig, J., Schneider, N., Schneider, L., Franke, U., Brox, T., Geiger, A.:
  {Sparsity invariant CNNs}. In: IEEE International Conference on 3D Vision
  (3DV). pp. 11--20 (2017)

\bibitem{van2019sparse}
Van~Gansbeke, W., Neven, D., De~Brabandere, B., Van~Gool, L.: {Sparse and noisy
  LiDAR completion with RGB guidance and uncertainty}. In: IEEE International
  Conference on Machine Vision Applications (MVA). pp.~1--6 (2019)

\bibitem{wang2009mean}
Wang, Z., Bovik, A.C.: {Mean squared error: Love it or leave it? A new look at
  signal fidelity measures}. IEEE Signal Processing Magazine  \textbf{26}(1),
  98--117 (2009)

\bibitem{16xu2019depth}
Xu, Y., Zhu, X., Shi, J., Zhang, G., Bao, H., Li, H.: Depth completion from
  sparse {LiDAR} data with depth-normal constraints. In: Proceedings of the
  IEEE/CVF International Conference on Computer Vision (ICCV). pp. 2811--2820
  (2019)

\bibitem{35xu2020deformable}
Xu, Z., Yin, H., Yao, J.: Deformable spatial propagation networks for depth
  completion. In: IEEE International Conference on Image Processing (ICIP). pp.
  913--917 (2020)

\bibitem{24yan2021rignet}
Yan, Z., Wang, K., Li, X., Zhang, Z., Xu, B., Li, J., Yang, J.: {RigNet}:
  Repetitive image guided network for depth completion. arXiv preprint
  arXiv:2107.13802  (2021)

\bibitem{yang2014color}
Yang, J., Ye, X., Li, K., Hou, C., Wang, Y.: Color-guided depth recovery from
  {RGB-D} data using an adaptive autoregressive model. IEEE Transactions on
  Image Processing  \textbf{23}(8),  3443--3458 (2014)

\bibitem{yang2019dense}
Yang, Y., Wong, A., Soatto, S.: Dense depth posterior (ddp) from single image
  and sparse range. In: Proceedings of the IEEE/CVF Conference on Computer
  Vision and Pattern Recognition (CVPR). pp. 3353--3362 (2019)

\bibitem{29zhao2021adaptive}
Zhao, S., Gong, M., Fu, H., Tao, D.: Adaptive context-aware multi-modal network
  for depth completion. IEEE Transactions on Image Processing  \textbf{30},
  5264--5276 (2021)

\bibitem{21zhong2019deep}
Zhong, Y., Wu, C.Y., You, S., Neumann, U.: Deep {RGB-D} canonical correlation
  analysis for sparse depth completion. Proceedings on the International
  Conference on Neural Information Processing Systems (NIPS)  \textbf{32}
  (2019)

\end{thebibliography}
\end{document}